\pdfoutput=1
\documentclass[11pt]{article}

\usepackage[]{EMNLP2022}

\usepackage{times}
\usepackage{latexsym}

\usepackage[T1]{fontenc}
\usepackage[utf8]{inputenc}

\usepackage{microtype}

\usepackage{inconsolata}

\usepackage{microtype}
\usepackage{amsthm}
\usepackage{amsmath}
\usepackage{amsfonts}
\usepackage{enumitem}
\usepackage{amssymb}
\usepackage{bbm}
\usepackage{algpseudocode}
\usepackage{booktabs}
\usepackage{tabularx}
\usepackage{amssymb}
\usepackage{xcolor,colortbl}
\usepackage{algorithm}
\usepackage{graphicx}

\definecolor{green}{rgb}{0.56, 0.74, 0.56}

\theoremstyle{definition}
\newtheorem{definition}{Definition}
\newtheorem{axiom}{Axiom}
\newtheorem{remark}{Remark}
\newtheorem{assumption}{Assumption}

\theoremstyle{plain}
\newtheorem{proposition}{Proposition}

\title{The Authenticity Gap in Human Evaluation}

\author{Kawin Ethayarajh \\
  Stanford University \\
  \texttt{kawin@stanford.edu} \\\ \And
  Dan Jurafsky \\
  Stanford University \\
  \texttt{jurafsky@stanford.edu}
  }

\begin{document}
\maketitle

\begin{abstract}
Human ratings are the gold standard in NLG evaluation.
The standard protocol is to collect ratings of generated text, average across annotators, and rank NLG systems by their average scores.
However, little consideration has been given as to whether this approach faithfully captures human preferences.
Analyzing this standard protocol through the lens of utility theory in economics, we identify the implicit assumptions it makes about annotators. 
These assumptions are often violated in practice, in which case annotator ratings cease to reflect their preferences.
The most egregious violations come from using Likert scales, which provably reverse the direction of the true preference in certain cases.
We suggest improvements to the standard protocol to make it more theoretically sound, but even in its improved form, it cannot be used to evaluate open-ended tasks like story generation.
For the latter, we propose a new human evaluation protocol called \textit{system-level probabilistic assessment} (SPA).
When human evaluation of stories is done with SPA, we can recover the ordering of GPT-3 models by size, with statistically significant results.
However, when human evaluation is done with the standard protocol, less than half of the expected preferences can be recovered (e.g., there is no significant difference between \texttt{curie} and \texttt{davinci}, despite using a highly powered test).
\end{abstract}

\section{Introduction}
\label{sec:introduction}

Human ratings are treated as the gold standard in NLG evaluation \citep{zhou2022deconstructing}.
For example, say one wants to claim that that their NLG model $X$ is better than the current state-of-the-art $Y$ and $Z$ for story generation.
The standard protocol is \textbf{outcome-level absolute assessment} (OAA): hire crowdworkers as annotators, collect individual ratings of a sample of stories generated by each model, and then claim that $X$ is the best because its average rating is the highest \citep{celikyilmaz2020evaluation}.
There is inconsistency in how this is implemented in the literature: terms such as `text quality' are often left undefined when instructing annotators \citep{howcroft2020twenty} and different papers use different rating scales \citep{amidei2019use}.
However, such criticism has been restricted to the implementations---little to no consideration has been given as to whether OAA can faithfully capture human preferences to begin with.

We start by analyzing the standard evaluation protocol through the lens of utility theory from economics (\S\ref{sec:reframing}).
We find that OAA can only capture an annotator's preferences under certain assumptions, which are unstated in the NLG literature and often violated in practice.
In such cases, annotator ratings become an unfaithful reflection of their preferences (\S\ref{sec:limitations}).
For example, by framing ratings as utility estimates, we extend a result from\phantom{.} \citet{boutilier2003foundations} to prove that using the same scale is insufficient for aggregating ratings \textit{across} annotators---they must agree on the maximum- and minimum-utility outcomes as well.
This precludes annotator ratings from being averaged unless they are given both maximally ``correct'' and ``incorrect'' references, which are available for some NLG tasks (e.g., machine translation) but not for open-ended ones (e.g, story generation), since the space of high-quality outputs is too diverse.
We provide concrete suggestions on how to improve the standard protocol to a point where it can faithfully capture human preferences in \textit{some} NLG tasks and settings (\S\ref{sec:implications}); however, for open-ended generation, an entirely new evaluation protocol is needed.

Though uncommon nowadays, a historic alternative to OAA was \textbf{outcome-level relative assessment} (ORA): create random pairs containing an output from $X$ and $Y$, ask annotators to pick the one they prefer in each, infer a score for $X$ and $Y$ that explains the results---based on a comparison model such as Bradley-Terry \citep{bradley1952}---and argue that $X$ is better because its estimated score is higher \citep{sakaguchi2014efficient}.
However, this also makes untenable assumptions about annotators; for example, even if $X$'s outputs are preferred to $Y$'s over 50\% of the time, $X$ may be less preferred to $Y$ if it has a tendency to fail catastrophically.
We observe that the main limitation of both OAA and ORA is their reliance on outcome-level judgments.

To this end, we propose \textbf{system-level probabilistic assessment} (SPA), which can be used for both open- and close-ended NLG tasks (\S\ref{sec:overcoming}).
SPA's key insight is that while an annotator cannot categorically determine whether they prefer system $X$ or $Y$---because the output space is too large for them to observe---they can estimate the \textit{probability} with which they prefer $X$ or $Y$ based on some fixed level of exposure to both.
SPA obviates assumptions about annotator preferences by delegating the responsibility of aggregating preferences over individual outputs into a preference over the underlying systems to the annotator themself, acknowledging that there is no canonical way to do so.
Because we are working with probabilities, aggregating across annotators is also straightforward.

We then ask annotators to use both the standard protocol (with a 5-point Likert scale) and SPA to express their preferences about the different GPT-3 variants w.r.t.\ their story-generation ability.
Given that larger GPT-3 variants generate more coherent, grammatical, and creative text, annotators\footnote{In aggregate, since individual annotators may have aberrant preferences.} should prefer each GPT-3 variant to the next smallest one, giving us 3 expected preferences \citep{brown2020language}.
Past work also suggests that annotators can distinguish human-written text from \texttt{ada} (the smallest variant) but not from \texttt{davinci} (the largest) \citep{clark2021all}, which gives us two additional expected preferences, for a total of 5.

When human evaluation is mediated by SPA, we can recover all 5 out of 5 expected preferences, with statistically significant results.
However, when human evaluation is done with the standard protocol, we can only recover 2 of the 5 preferences, despite all tests having statistical power $\approx 1$ for $\alpha = 0.001$.
The standard protocol also yields the surprising---and likely incorrect---result that human-written text is significantly \textit{less} preferred to \texttt{davinci}.
The failures of the standard protocol suggest that its theoretical limitations have practical consequences, and the flexibility of SPA makes it the better option in most intrinsic evaluation settings.

\section{Reframing Human Evaluation}
\label{sec:reframing}

To understand what causes human preferences to be misrepresented, we will analyze NLG evaluation through the lens of economic theory on preference modeling.
In doing so, we find that comparing NLG systems is an instance of a common problem in utility theory.
To begin, let $X,Y$ denote the NLG systems to be compared and annotator $a_i$ be the \textit{agent} making the comparisons.
As we are drawing from the economics literature, we will primarily use economic terms such as \textit{lottery} and \textit{utility} in our framing, which we will define as we go along.

\subsection{NLG Systems as Lotteries}

\begin{definition}[\textbf{Lottery}]
A \textit{lottery} is a probability distribution over a space of finite outcomes \citep{boutilier2003foundations}. Given a (possibly empty) prompt or input, an NLG system induces a lottery over all possible output text.
\end{definition}

Given that there is also a discrete distribution over the prompts/inputs used, the lottery that the NLG system induces over the output text is itself the outcome of a lottery over the prompts/inputs. 
This means that $X$ and $Y$ are \textit{compound lotteries}: a lottery of a lottery, which can be reduced to a simple lottery over the output text by marginalization. Thus for a known prior over the prompts/inputs, we can think of any NLG system as a simple lottery over all possible output text.

\subsection{Choices as Preference Relations}

\begin{definition}[\textbf{Preference Relations}]
The \textit{relation} $X \succ_i Y$ means that the agent $a_i$ strictly prefers $X$ to $Y$; the relation $X \prec_i Y$ means that the agent strictly prefers $Y$ to $X$; the relation $X \sim_i Y$ means that the agent is indifferent to the two. Relations without the subscript $i$ denote the aggregate preference across all agents.
\end{definition}

This means that determining whether an annotator prefers one NLG system to another is an instance of a common problem in economics: determining which of two lotteries the agent prefers.
For most such problems in the real world, we could ask the agent to directly compare the two lotteries (e.g., we could ask an investor what split of stocks to bonds they would invest in) \citep{mankiw2020principles}.
However, because in NLG the output space is so large, we cannot ask an annotator to categorically determine which of two lotteries they prefer.
What is feasible is asking an annotator to compare two individual output texts, but there is no assumption-free means of aggregating preferences over individual outcomes into preferences over the lotteries (\S\ref{ssec:assumptions}).

\subsection{Text Quality as Utility}
\label{ssec:text_quality}

\begin{definition}[\textbf{Utility}]
Abstractly, the \textit{utility} of a good denotes the benefit that an agent receives from it. The \textit{utility function} $u_i: S \to \mathbb{R}$ for agent $a_i$ maps outcomes $S$ to real values based on the utility derived \citep{mankiw2020principles}.
For NLG, the utility of a text is how good the agent perceives it to be, optionally w.r.t.\ some attribute such as coherence. 
\end{definition}
    
\begin{definition}[\textbf{Ordinal Utility}]
Under the theory of \textit{ordinal utility}, only the ranking induced by $u_i$ matters; the magnitude of the difference between the values do not \citep{mankiw2020principles}. An ordinal utility function $u_i$ \textit{represents} $\succ_i$ if it preserves the ranking the latter induces: $ X \succeq_i Y \iff u_i(X) \geq u_i(Y)$.
Two utility functions $u,v$ are \textit{ordinally equivalent} if they induce the same preference ordering.
\end{definition}

\begin{definition}[\textbf{Cardinal Utility}]
Under the theory of \textit{cardinal utility}, the magnitude of the difference between two outcomes' utility does matter. Two utility functions $f,g$ are \textit{cardinally equivalent} up to a positive affine transformation \citep{dybvig1981recovering}.
\end{definition}

Estimating cardinal utility is the approach that has been implicitly taken by the standard evaluation protocol for NLG (a.k.a., outcome-level absolute assessment (OAA)).
When an annotator rates an example, they are estimating the cardinal utility $u_i(x)$ they get from an outcome $x$.
When those ratings are averaged to score the system $X$ that produced those examples, one is estimating the expected utility of a lottery.
Estimating the cardinal utility of a lottery as the expected utility of its outcomes is possible because of the von Neumann-Morgenstern theorem \citep{morgenstern1953theory}.
No similar result exists for estimating the ordinal utility, however---we cannot average rankings.

\subsection{Assumptions of Agent Rationality}
\label{ssec:assumptions}

Outcome-level relative assessment (ORA) explicitly encodes its assumptions about annotator preferences in a comparison model such as Bradley-Terry \citep{bradley1952} or Thurstone \citep{thurstone1927law}.
These assumptions are easy to identify and invalidate, so we refer the reader to prior work on its limitations \citep{sakaguchi2014efficient,bojar2016findings}.
ORA has also declined in popularity in recent years, with OAA now making up a supermajority of human evaluation \citep{van2021human}.
Because it does not use a comparison model, the now widely-used OAA may seem as though it makes no assumption about annotators.
However, ranking systems by their average rating only captures annotator preferences if they are Von Neumann-Morgenstern-rational agents \citep{morgenstern1953theory}:

\begin{definition}[\textbf{VNM Rationality}]
\label{thm:vnm-rationality}
    Let $X',  Y'$ denote random variables representing the outcomes of lottery $X, Y$ respectively. 
    For any \textit{von Neumann-Morgenstern-rational} agent, there exists a utility function $u_i$ such that $X \succeq_i Y \iff \mathbb{E}[u_i(X')] \geq \mathbb{E}[u_i(Y')]$. 
    In other words, VNM-rational agents always choose to maximize their expected utility. 
    In order for an annotator $a_i$ to be a VNM-rational agent, their preferences must satisfy the following four axioms for any NLG systems $X, Y, Z$:
    
    \begin{axiom}[\textbf{Completeness}]
        For any $X, Y$, exactly one of the following holds for each agent $a_i$: $X \succ_i Y, X \prec_i Y$ or $X \sim_i Y$ (i.e., the agent prefers $X$, prefers $Y$, or is indifferent respectively).
    \end{axiom}

    \begin{axiom}[\textbf{Transitivity}]
    If $X \succeq_i Y$ and $Y \succeq_i Z$, then $X \succeq_i Z$.
    \end{axiom}
    
    \begin{axiom}[\textbf{Continuity}]
    If $X \succeq_i Y \succeq_i Z, \exists\ p \in [0,1]$ such that $p X + (1 - p)Z \sim_i Y$.
    \end{axiom}
    
    \begin{axiom}[\textbf{Independence}]
    \label{axiom:independence}
    For any $Z$ and $p \in (0,1]$, we have $X \succeq_i Y \iff pX + (1 - p)Z \succeq_i Y + (1 - p)Z$. 
    \end{axiom}

\end{definition}
Although it may seem intuitive that any agent would maximize their expected utility, work in behavioral economics has identified many situations where agents choose not to do so \citep{samuelson1977st,kahneman1979prospect,allais1979so}.

\section{Causes of Misrepresentation}
\label{sec:limitations}

By framing human evaluation in terms of utility theory, we found that the standard protocol in NLG evaluation serves to estimate the cardinal utility of a system via outcome-level absolute assessment (\S\ref{ssec:text_quality}).
We then listed the assumptions that agent preferences need to satisfy in order to make this estimation valid (\S\ref{ssec:assumptions}).
In this section, we discuss how these assumptions are often violated in NLG evaluation, and how this begets misrepresentation of an annotator's true preferences.
We limit our criticism to OAA in this section, since ORA has already been criticized in prior work \citep{sakaguchi2014efficient} and has, over the past several years, become far less common than OAA \citep{van2021human}.

We begin by noting that rating generated text is done one of two ways \citep{celikyilmaz2020evaluation}:
\begin{enumerate}[leftmargin=*]
    \item Likert scales\footnote{To be more specific, a Likert scale is a collection of Likert items, each of which is a discrete rating from 1-to-$k$.}, which discretize the utility into an integer from 1-to-$k$, usually 1-to-5.
    \item Continuous scales (a.k.a., continuous direct assessment), which normalize the utility from 0-to-$k$ ($k$ usually being 100).
\end{enumerate}
Our first critiques apply only to Likert scales, but our last two apply to the standard protocol at-large.

\begin{remark}[\textbf{Ordinal-Cardinal Conflation}]
Averaging ordinal Likert ratings to estimate cardinal utility can violate tenets of utility theory.
\end{remark}

The Likert scale is ordinal: an outcome with a higher score is preferred to one with a lower score, but the distance between the points is not significant.
In contrast, the intervals \textit{are} significant in cardinal utility.
Averaging Likert ratings to estimate cardinal utility thus assumes that the annotator has perceived the distance between each point to be the same, which is impossible to verify.
At best, annotators can be steered into an interval-based interpretation through careful wording of the question, but there is no guarantee that they will interpret the distances as intended.
In a survey of the NLG literature, \citet{amidei2019use} found that 31 of 38 papers using Likert scales took an interval-based interpretation of them, but only 1 paper provided justification for this interpretation.

This problem is not solved by normalization methods such as $z$-scoring, as they do not work when the interval widths are asymmetric (e.g., the annotator might perceive the jump between 1-to-2 to be larger than the jump from 2-to-3 on a 3-point scale).
This is not a novel observation either; there is extensive work on the limitations of averaging over Likert ratings \citep{jamieson2004likert,sullivan2013analyzing,pimentel2019some}.
Even early shared tasks for NLG expressed this concern and used continuous scales instead \cite{gatt2009introducing}.

\begin{remark}[\textbf{Biased Estimation}]
Averaging Likert ratings can be a biased estimator of the expected utility, potentially reversing the direction of the true preference over two NLG systems.
\end{remark}

Building upon Remark 1, let us make a best-case assumption that the annotator perceives the intervals between the points on the Likert scale to be equal. 
As such, they determine the Likert score by normalizing their utility to [0,5] and then applying the ceiling function (e.g., $[0,1] \to 1; (1,2] \to 2$, etc.).\footnote{Using a window of 0.5 around each number and rounding would make the 1-star bucket larger than the rest.}
This effectively replaces a subset of preference relations $\succ_i$ with indifference relations.
That is, if two texts both have utilities in the tier $(i, i+1]$, the annotator becomes indifferent to them because of this transformation.

This can be stated more generally:

\begin{proposition}
Let $r_i(s) := \lceil u_i(s) \rceil - u_i(s)$. Without loss of generality, if $\mathbb{E}_{s \sim X}[r_i] > \mathbb{E}_{s \sim Y}[r_i]$, then Likert ratings over-estimate the utility of lottery $X$ relative to $Y$; if $\mathbb{E}_X[r_i] < \mathbb{E}_Y[r_i]$, they under-estimate the utility of $X$ relative to $Y$.
\end{proposition}

\begin{proposition}
Let $\mathbb{E}[u_i(X')] > \mathbb{E}[u_i(Y')]$ without loss of generality. If $(\mathbb{E}[u_i(X')] - \mathbb{E}[u_i(Y)']) < (\mathbb{E}_Y[r_i] -\mathbb{E}_X[r_i])$, then averaging Likert ratings reverses the direction of the true preference.
\end{proposition}

Since our annotator is implicitly assumed to be VNM-rational, by the von Neumann-Morgenstern theorem, $X \succ_i Y \iff \mathbb{E}[u_i(X')] > \mathbb{E}[u_i(Y')]$.
Including the residuals can potentially change the direction of the inequality between the expected utilities.
Thus by the VNM theorem, it can also change the direction of the preference relation.
Since $r \in [0,1]$, the difference $|\mathbb{E}_Y[r_i] -\mathbb{E}_X[r_i]| \leq 1$, meaning that a reversal of preference could only occur when the annotator perceived both NLG systems to produce outcomes of similar utility on average.
This is a common situation in practice, as proposed systems are often an incremental improvement over the state-of-the-art \citep{card2020little}.

\begin{remark}[\textbf{Non-Independent Lotteries}]
Lottery independence is an axiom of VNM-rationality but often fails to hold in practice for NLG systems.
\end{remark}

One of the conditions that needs to be satisfied for VNM-rationality is independence over lotteries, as defined in Axiom \ref{axiom:independence}.
Put simply, the preference $X \succ_i Y$ should not change if another lottery $Z$ is mixed with both in equal proportion.
However, this assumption is often violated in the real world.
Say that $X, Y$ place zero mass on offensive text (e.g., swear words).
This is typical for consumer-facing NLG systems, which may explicitly filter out such outputs to avoid public outcry, the loss of users, and a potential lawsuit \citep{zhou2022deconstructing}.
If lottery $Z$ places any mass on offensive output, adding it to either $X$ or $Y$ may result in the system being unusable. 
If both systems become unusable, the relation between the lotteries would change from preference ($X \succ_i Y$) to indifference ($X \sim_i Y$), despite the direction of the
expected utility inequality remaining the same.
In such a case, the agent would not be VNM-rational, meaning that their preference could not be inferred by comparing the expected utility of each NLG system.

\begin{remark}[\textbf{Inter-Agent Incomparability}]
Using the same scale across annotators is insufficient for aggregating their cardinal utility (i.e., estimating the \textit{expected expected utility}) due to differences in the magnitude of utility.
\end{remark}

When ranking NLG systems, we do not want to rank them according to just one individual, since that individual's preferences may be unrepresentative of the population.
In other words, there is a distribution over utility functions, and we want to estimate the expected utility w.r.t.\ this distribution.
This quantity is also known as the expected expected utility (EEU):  $\mathbb{E}_i [ \mathbb{E}[u_i(X')]]$ \citep{boutilier2003foundations}, which can be expanded as
\begin{equation}
\label{eq:eeu}
     \text{EEU}[X] = \int \mathbb{E}[u_i(X')] p(u_i) du_i
\end{equation}
Then we could infer the direction of the aggregate preference over the entire agent population, since $X \succ Y \iff \text{EEU}[X] > \text{EEU}[Y]$.

Estimating the EEU is not as straightforward as averaging the expected utility estimates of different agents.
Given a continuous scale from 0-to-100, one agent may score in the range 0-to-10 while another may score in 90-to-100.
Averaging across the two agents would bias the one with a greater magnitude of scoring.
In technical terms, EEU is not invariant to the choice of utility function in a set of cardinally equivalent utility functions.
This has been observed empirically in NLP and been framed as annotators being too strict or too forgiving \citep{zemlyanskiy2018aiming,kulikov2019importance}.

Presenting all annotators with the same scale does not necessarily solve this problem, since it does not force annotators to adopt the same magnitudes.
$Z$-scoring does not necessarily solve this problem either, since the annotator scores are not guaranteed to be normally distributed.
Relative magnitude estimation \citep{moskowitz1977magnitude,novikova2018rankme}, where the annotator provides the score of an outcome relative to some reference, \textit{partially} addresses this problem, but using a single arbitrary reference point is not provably sufficient.

\citet{boutilier2003foundations} formally proved that in addition to the continuity axiom (\S\ref{ssec:assumptions}), \textit{extremum equivalence} is sufficient to estimate EEU, which he defined as: (1) all agents agree on the most and least preferred outcomes; (2) all agents assign their most and least preferred outcomes the utility $c_\text{max}, c_\text{min}$ respectively, where $c_\text{max} > c_\text{min} \geq 0$.
These conditions might be satisfied in machine translation, for example; one could argue that providing ``correct'' and ``incorrect'' references forces all annotators to share utility function endpoints.
But when there are no references or the space of high-quality outputs is diverse, as in open-ended NLG tasks (e.g., chitchat dialogue), this condition cannot be satisfied.
% Note that extremum equivalence is a sufficient but not necessary condition; it is possible that there are more relaxed conditions under which averaging expected utility estimates is viable.
% However, the NLG literature does not consider any such conditions and treats averaging over annotators as a given.

\section{Improving the Standard Protocol}
\label{sec:implications}

By making some minor changes, the OAA-based standard evaluation protocol can be improved to a point where it can adequately capture human preferences in \textit{some} NLG tasks and settings:

\begin{enumerate}[leftmargin=*]

    \item Continuous scales should be used instead of Likert scales to avoid ordinal-cardinal conflation and potentially biased estimation.
    % This need was pointed out even in early shared tasks for NLG \citep{gatt2009introducing}.
    
    \item To satisfy extremum equivalence (\S\ref{sec:limitations}, Remark 4), both maximal- and minimal-utility references should be provided, effectively forcing all annotators' utility functions to share endpoints.
    This can only be done when the space of ideal outcomes for a given input is small and well-defined (e.g., machine translation).
    % Doing so can manipulate agents' utility functions, which may be desired when evaluating w.r.t.\ a particular attribute such as fluency, but may not be desired when trying to capture true human preferences.
    % If extremum equivalence cannot be enforced, only one annotator should be used for all ratings, though this will yield ungeneralizable conclusions.
        
    \item To satisfy lottery independence, there should be no outcome that can make an NLG system unusable (e.g., because the system is only used by a limited set of users whose utility is bounded).
    
\end{enumerate}
The WMT competition for machine translation---which has experimented with many evaluation schemes---has had, since 2017, a protocol that follows many of these suggestions \citep{bojar2017findings,specia2021findings}.
It uses continuous scales, provides maximum-utility references, and hires translators, meaning lottery independence is safe to assume.
Still, this improved protocol cannot be applied to open-ended tasks where there is no singular notion of correctness, tasks where maximal-utility outcomes can be diverse (e.g., story generation), or when lottery independence is likely to be violated in the real-world (e.g., offensive chatbots).
Such tasks and settings demand an entirely new evaluation protocol (\S\ref{sec:overcoming}).
\section{System-level Probabilistic Assessment}
\label{sec:overcoming}

The limitations of both ORA and OAA stem from trying to aggregate preferences over outcomes into a preference over systems, despite there being no canonical way to do so.
For example, one annotator may prefer $X$ to $Y$ only if the former wins head-to-head comparisons of outputs over 50\% of the time, but another annotator may choose by comparing the worst-case output from each system.
Therefore we propose directly asking annotators to estimate the probability $P[X \succ_i Y]$ that a preference holds across two systems, a protocol we call \textit{system-level probabilistic assessment} (SPA).

\subsection{Theory}
\label{ssec:spa_theory}

Let $P[X \succ Y]$ denote the aggregate preference probability of $X \succ Y$ for a population of agents.
Where $p(\succ_i)$ is the frequency of preferences $\succ_i$, we can expand $P[X \succ Y]$ similarly to EEU (\ref{eq:eeu}):
\begin{equation}
    P[X \succ Y] = \int P[X \succ_i Y] p(\succ_i) d\succ_i
\end{equation}
Since $P[X \succ_i Y] \in [0,1]$ for all $a_i$, the values are inherently comparable across annotators, making inter-annotator aggregation easy.
As in comparison models \citep{bradley1952}, this one measure is sufficient to infer the direction of the preference: annotators are indifferent iff $P[X \succ Y] = 0.5$ (i.e., no different than chance); $X$ is more(less) preferred to $Y$ if $P[X \succ Y]$ is greater(less) than 0.5. 
In practice, however, statistical significance is important to consider (see \S\ref{ssec:new_protocol} for details).

If we assumed preferences were complete, then $P[X \succ_i Y]$ could only take a value in $\{0,1\}$, but doing so would be unrealistic, since annotators are almost never exposed to the entirety of an NLG system's output in practice, precluding them from preferring one system with absolute certainty.
Therefore we model preferences as stochastic.
Modeling preferences as stochastic is not new \citep{bradley1952,thurstone1927law}, but the approach has traditionally been to use categorical preference labels to learn stochastic models \citep{chu2005preference}.
The novelty of our approach is that we ask the annotators themselves to estimate the preference probability.

However, an annotator's preferences change as they are exposed to more output while $P[X \succ_i Y]$ is a fixed value.
How can we reconcile this?
Every time an annotator's preference probability is updated, they effectively become a new agent (i.e., an agent is not an individual annotator, but a specific iteration of an annotator with fixed beliefs).
For example, at the start, an annotator has no knowledge of the systems, so $P[X \succ_{i,t = 0} Y] = 0.5$.
As they are exposed to more outputs, they may develop a preference for one system (e.g., $P[X \succ_{i,t = 1} Y] = 0.7$).
At some point they will become certain about their choice (e.g., $P[X \succ_{i,t = \infty} Y] = 1$), but at this point the annotator is no longer the same agent that was split between the two options.
In other words, agent $a_i$ is uniquely defined by an annotator $a$ and their level of exposure $t$.

The standard protocol in NLG evaluation requires that annotators be VNM-rational and have preferences that are complete, transitive, independent, continuous, and extremum equivalent (\S\ref{ssec:assumptions}).
SPA obviates those assumptions by delegating the responsibility of aggregating preferences over outcomes into a preference over the underlying lotteries to the agent themself, acknowledging that there is no canonical way to do so.
Estimating $P[X \succ Y]$ only requires two assumptions:

\begin{assumption}[\textbf{Unbiased Stated Preferences}]
An agent $a_i$ has unbiased stated preferences if, when asked to estimate the probability of their preference for lottery $X$ over $Y$, they provide an unbiased estimate $\hat{P}[X \succ_i Y]$ (i.e., the noise has expectation zero).
\end{assumption}

\begin{assumption}[\textbf{Complementarity and Indifference}]
$P[X \succ_i Y] + P[Y \succ_i X] = 1$, and $X \sim_i Y \iff P[X \succ_i Y] = P[X \prec_i Y] = 0.5$. 
% (i.e., an agent is indifferent if and only if the probability of preferring a system is no different from chance).
\end{assumption}

\subsection{Implementing SPA}
\label{ssec:new_protocol}
If you want to use SPA to compare two NLG systems $X,Y$, you should do as follows:

\begin{enumerate}[leftmargin=*]
    \item Find $n_A$ \underline{unique} annotators who are representative of the agent population whose preferences you want to model.
    Choose the prior for your desired task and draw $m$ prompts/inputs from this prior.
    Give each annotator $m$ randomly sampled outputs from each system, one per prompt.
    It is possible to use a different set of prompts for $X$ and $Y$, but this will make it harder for the agent to do an apples-to-apples comparison, making them less certain about their preference.
    
    \item Ask each annotator a variation of the question: ``Based on what you've read, from 0 to 100, what is the \% chance that system $X$ is a better writer than system $Y$?'' 
    Swapping $X$ with $Y$ and then asking the question will not necessarily equal $1 - \hat{P}[X \succ_i Y]$, since the estimates are noisy. 
    
    \item (\textbf{optional}) To filter out annotators with a poor understanding of probability, ask annotators to estimate both $\hat{P}[X \succ_i Y]$ and $\hat{P}[Y \succ_i X]$ and exclude those for whom $\hat{P}[X \succ_i Y] + \hat{P}[Y \succ_i X] > \tau = 1.1$.
    We set $\tau \gets 1.1$ instead of 1.0 to account for noisy estimates. 
    If multiple systems are being compared, exclude annotators that fail this condition even once.
    
    \item Estimate the aggregate probability $P[X \succ Y]$ by averaging over $\{ \hat{P}[X \succ_i Y] \}$.
    Use a two-sided Student's $t$-test to determine whether it is significantly different from chance (0.5).
    If $P[X \succ Y]$ is significantly higher(lower) than 0.5 at level $\alpha$, then you can reject the null hypothesis that $X \sim Y$ at significance level $\alpha$.
    If $P[X \succ Y]$ is not significantly different from 0.5, then the null hypothesis that $X \sim Y$ cannot be rejected.
\end{enumerate}
In the Appendix, we provide details of the SPA implementation we use in our experiments in \S\ref{ssec:experiments}.

\begin{table*}[t]
\begin{tabularx}{\textwidth}{Xlccc}
\toprule
System $X$ & System $Y$ & Expected Preference & $P[X \succ Y]$ (SPA) &  Likert Rating $\Delta$  \\
\midrule
$\texttt{GPT-3-ada}$ & $\texttt{human}$ & $X\prec Y$ &   \cellcolor{green} 0.420*\ \ \ \ \ &       \cellcolor{green} $-$0.822*** \\

$\texttt{GPT-3-babbage}$ & $\texttt{GPT-3-ada}$ & $X \succ Y$ &   \cellcolor{green} 0.688*** &         \ \ \ \cellcolor{green} 0.644*** \\

$\texttt{GPT-3-curie}$ & $\texttt{GPT-3-babbage}$ & $X \succ Y$ &         \cellcolor{green} 0.630*** &         \ \ \ 0.322\ \ \ \ \ \ \    \\

$\texttt{GPT-3-davinci}$ & $\texttt{GPT-3-curie}$ &  $X \succ Y$ &        \cellcolor{green} 0.575*** &        \ \ \  0.244\ \ \ \ \ \ \    \\

$\texttt{human}$ & $\texttt{GPT-3-davinci}$ & $X \sim Y$ &        \cellcolor{green} 0.544\ \ \ \ \ \ \   & \cellcolor{red} $-$0.389* \ \ \ \ \     \\
\bottomrule
\end{tabularx}
\caption{Eliciting preferences for story generation, using both system-level probabilistic assessment (SPA) and the standard protocol with 5-point Likert ratings.
We use two-sided Student's $t$-tests with Holm-Bonferroni-corrected significance at $\alpha = 0.10 (^{*}), 0.05 (^{**}), 0.01 (^{***})$. 
SPA consistently yields a significant result in the expected direction; the standard protocol, only twice (green).
Notably, the latter suggests that human-written text is significantly \textit{less} preferred to $\texttt{davinci}$-written text (red), although past work has found that annotators cannot tell the difference \citep{clark2021all}. 
SPA finds $P[\texttt{human} \succ \texttt{davinci}]$ to not be significantly different from chance.
}
\label{tab:table}
\end{table*}

\section{Experiments}
\label{ssec:experiments}

\subsection{GPT-3 Story Generation}
\label{ssec:gpt3}

To test our proposed protocol, we ask 90 {unique} crowdworkers to use both the standard protocol (with a 5-point Likert scale) and SPA to express their preferences about the different GPT-3 variants w.r.t.\ their story-writing ability (see Appendix \ref{sec:gpt3_appendix} for details).
The story prompts are drawn from the WritingPrompts dataset \citep{fan2018hierarchical} and each annotator is given: $m$ randomly drawn prompts, stories generated by each GPT-3 variant for those prompts, and a human-written story for each prompt.
The annotator is not told which of the 5 systems is a human.
With SPA, they are asked to compare the systems themselves, while with the standard protocol, they are just asked to rate the outputs.
The smaller $m$ is, the more uncertain annotators will be about their preference, making it hard to elicit a statistically significant result in SPA.
The larger $m$ is, the higher the per annotator cost, since the task will take longer to complete.
We balance these concerns by choosing $m = 5$.

Given that larger GPT-3 variants generate more coherent and creative text, annotators in aggregate should prefer larger variants: i.e., $\texttt{davinci} \succ \texttt{curie} \succ \texttt{babbage} \succ \texttt{ada}$ \citep{brown2020language}.
\citet{clark2021all} also found that annotators can distinguish between GPT2- and human-written text, but not at all between human- and \texttt{davinci}-written text.
Since \texttt{ada} is not much larger than GPT2, this implies that the following preferences should also hold: $\texttt{human} \succ \texttt{ada}$ and $\texttt{human} \sim \texttt{davinci}$.
For SPA, we use a two-sided Student's $t$-test to measure whether each probabilistic preference is significantly different from chance ($P[X \succ Y] = 0.5$).
For the standard protocol, we use a paired $t$-test to determine whether the Likert ratings of two systems' outputs are significantly different.
Since we make multiple comparisons, we apply the Holm-Bonferroni correction \citep{holm1979simple}.

As seen in Table \ref{tab:table}, SPA recovers 5/5 expected preferences: each GPT-3 variant is significantly preferred to the next smallest one; $\texttt{ada}$ is significantly less preferred to human-written text; and annotators are indifferent to human- and \texttt{davinci}-written text.
However, the standard protocol only recovers 2/5 expected preferences: \texttt{curie} and \texttt{babbage} are not significantly preferred to the next smallest GPT-3 variant, and the human writer is significantly \textit{less} preferred to \texttt{davinci}, despite past work suggesting that annotators cannot tell the difference between the two \citep{clark2021all}.

\subsection{DALL-E Image Generation}
\label{ssec:dalle}

Though the focus of this work is NLG evaluation, SPA can also be used to evaluate other types of generated content.
We run a similar experiment with image generation, where the system \texttt{DALL-E-scrambled} scrambles the prompt before feeding it to DALL-E\footnote{https://openai.com/blog/dall-e/} and the system \texttt{DALL-E-raw} does not (see Appendix \ref{sec:dalle_appendix} for details).
For example, given the prompt \textit{`ball on a chair'}, \texttt{DALL-E-raw} feeds the prompt as is to DALL-E while \texttt{DALL-E-scrambled} may feed it \textit{`chair on a ball'}.
Annotators are then asked which system's images are better, where the goodness of an image is defined as how interesting, coherent, and relevant it is to the original (unscrambled) prompt.
Given that scrambling the text affects its compositionality, images generated using the scrambled prompt should be no better than those generated with the original prompt; we expect to recover $\texttt{DALL-E-raw} \succ \texttt{DALL-E-scrambled}$.

\begin{figure}
    \centering
    \includegraphics[width=\columnwidth]{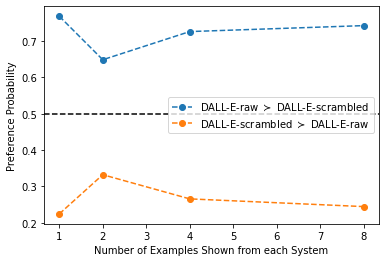}
    \caption{According to SPA, scrambling the prompt before using DALL-E creates significantly worse images than using the original prompt, as expected ($\alpha=0.01$, Holm-Bonferroni-corrected).
    The conclusion is the same regardless of whether we ask agents to estimate $P[\texttt{DALL-E-raw} \succ \texttt{DALL-E-scrambled}]$ or vice-versa.
    }
    \label{fig:design}
\end{figure}

Our goal with this experiment is to understand how design choices may affect the conclusion drawn with SPA.
For one, as shown in Figure \ref{fig:design}, we find that increasing $m$, the number of examples shown to the agent, makes them more certain in their preference, pushing the probability to $0$ or $1$.
But there are diminishing returns: the jump from $4 \to 8$ examples yields less of a change in the preference probability than from $2 \to 4$.
Note the outlier at $m = 1$: surprisingly, agents are quite certain about which system is better when they only see one example from each.
We believe that this over-confidence stems from failing to imagine the variance in image quality produced by a single system, variance that the agent is exposed to at $m > 1$.
This reaffirms our choice of $m = 5$ in \S\ref{ssec:gpt3}, and we encourage readers to use $m \geq 4$ in practice.

Secondly, it should not matter whether we ask annotators to estimate $P[X \succ_i Y]$ or $P[Y \succ_i X]$, since by Assumption 2, $P[X \succ_i Y] = 1 - P[Y \succ_i X]$.
As seen in Figure \ref{fig:design}, this holds in practice; for all values of $m$, the absolute distance of $P[\texttt{DALL-E-scrambled} \succ \texttt{DALL-E-raw}]$ and $P[\texttt{DALL-E-raw} \succ \texttt{DALL-E-scrambled}]$ from 0.5---a measure of the annotators' conviction in their preference---is not significantly different.
This is important, as it implies that practitioners cannot ``hack'' SPA by changing the ordering of the systems in the question posed to the annotators.

\subsection{Discussion}

\begin{figure}
    \centering
    \includegraphics[width=\columnwidth]{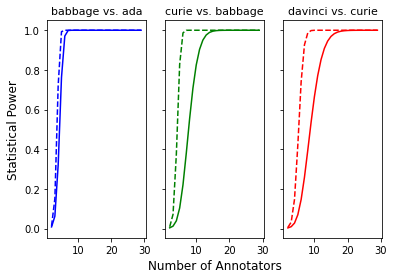}
    \caption{The statistical power of our experiments as a function of the number of annotators $n_A$, both for SPA (dashed) and the standard protocol (solid), assuming the observed effect size stays constant and $\alpha = 0.001$. The observed differences are unlikely to be explained simply by the larger sampling variance of one protocol: at the observed effect sizes, both tests would have high power. }
    \label{fig:stat_power}
\end{figure}

Why does SPA work better than the standard protocol, successfully recovering $\texttt{curie} \succ \texttt{babbage}$ and $\texttt{babbage} \succ \texttt{ada}$ while the latter does not?
In Figure \ref{fig:stat_power}, we show that this is \textit{not} due to statistical power; since we use $n_A = 90$ (after excluding the 10 annotators that did not follow instructions), the power of all our statistical tests---both using SPA and the standard protocol---is $\approx 1$ for $\alpha=0.001$.
Given the sample size and observed variability, the failure of the standard protocol does not appear to be explained simply by insufficient precision.
We contend that this is because the elicited Likert ratings do not represent the annotators' true preferences to begin with (\S\ref{sec:limitations}).
Just as replacing annotator judgments with random noise would preclude us from rejecting the null hypothesis, the judgments collected via the standard protocol are so distorted that a highly powerful test fails to recover the anticipated preference.

\section{Related Work}
\label{sec:related}

Soliciting humans to directly evaluate the quality of generated text is known as \textit{intrinsic evaluation}.
The text can be judged for its overall quality or along a specific dimension such as coherence, though these terms are not consistently defined \citep{howcroft2020twenty,van2021human}. 
This is most often done in the NLG literature by having annotators assign a Likert rating from 1-to-$k$, where $k$ is usually 5 \citep{van2019best}.
Given the ubiquity of this standard protocol, little justification is given when it is used and implicit assumptions, such as equal intervals for Likert scales, are entirely omitted \citep{amidei2019use}.

The earliest shared tasks in NLG, such as the \texttt{TUNA} \citep{gatt2009introducing} and \texttt{GREC} \citep{belz2011discrete} challenges for expression generation, used a continuous scale for scoring, explicitly noting that annotators may not perceive the intervals on a Likert scale to be equal.
In contrast, early modeling work---such as the \texttt{STORYBOOK} system for narrative prose generation \citep{callaway2002narrative}---used discrete ratings.
This difference in evaluation protocol between shared challenges and individual modeling papers continued over the years.
For example, the E2E NLG challenge \citep{duvsek2018findings} used continuous scores based on relative magnitude estimation \citep{novikova2018rankme,bard1996magnitude}.
However, these challenges have not served as a bulwark against the popularity of Likert-based OAA. Even recent attempts to standardize human evaluation in NLG---using evaluation platforms---collect Likert ratings \citep{khashabi2021genie,gehrmann2021gem}.

Compared to OAA, outcome-level relative assessment (ORA) is far less common nowadays, in large part because the cost of pairwise output comparisons grows combinatorially as you evaluate more systems \citep{celikyilmaz2020evaluation}.
Recall that given binary outcome-level preferences (e.g., $x_i \succ y_i$) as labels, ORA uses a preference model such as Bradley-Terry to estimate the scores of the systems, analogous to how ELO scores are calculated for chess players \citep{chu2005preference}.
In explicitly stating its assumptions about annotator preferences using a preference model, ORA was easier to criticize than OAA, which contributed to the former's decline \citep{sakaguchi2014efficient}.
The one area in which comparison-based evaluation still prevails is when conducting a Turing test---seeing whether annotators do better than chance when guessing whether a text is human- or machine-generated \citep{garbacea2019judge,ippolito2020automatic,brown2020language,clark2021all}.
This is acceptable, since what is being measured is not annotator preference but rather discriminability.

Over the years, machine translation (MT) has had spirited debate about evaluation.
\citet{callison2007meta} found that compared to ranking outputs, annotators took more time and agreed less when providing Likert scores.
Citing this, \citet{sakaguchi2014efficient} use the TrueSkill algorithm \citep{herbrich2006trueskill} to estimate scores for NLG systems based on pairwise preferences of their output.
This approach, called \emph{relative ranking} (RR) was used in the WMT competition until 2016, when \textit{direct assessment} (DA) on a 0-to-100 continuous scale were trialled and found to produce systems rankings that strongly correlated with RR \citep{bojar2016findings}.
DA also had the advantage of providing an absolute measure of quality, so it was adopted as the standard for WMT in 2017 and used thereafter \citep{bojar2017findings,specia2021findings}.

% To our knowledge, utility theory has only been applied in NLP to design leaderboards \citep{ethayarajh2020utility,ma2021dynaboard}.
% However, past work has hinted at utilitarian issues without framing them as such, such as whether crowdworkers are sufficiently motivated \citep{belz2006comparing,dugan2020roft,mitra2015comparing} and aligning annotators' incentives with that of potential users \citep{oppenheimer2009instructional,daniel2018quality}.
\section{Conclusion}

We analyzed the standard evaluation protocol in NLG through the lens of utility theory, finding that it makes untenable assumptions about annotator preferences.
When these assumptions are violated, annotator ratings become an unfaithful reflection of their preferences, both in theory and in practice.
We proposed a new evaluation protocol called SPA that makes minimal assumptions---not only is it more theoretically sound than the standard protocol, but it performs better in practice as well, consistently recovering the expected preference with statistically significant results.
An important direction of future work will be re-evaluating conclusions in the NLG literature with SPA and seeing which ones stand up to scrutiny.

\section*{Limitations}
Although SPA does not suffer from the existential limitations of the standard evaluation protocol (\S\ref{sec:limitations}), it does have three notable limitations.

\begin{enumerate}[leftmargin=*]
    \item SPA does not measure the magnitude of a preference, only the probability that it exists.
    The magnitude of a preference is useful for understanding the trade-offs involved in deploying one NLG system over another---even if a new system is certainly more preferred to an older one, it might not be worth deploying if the magnitude of the preference is small.
    This is a necessary trade-off for SPA to be applicable to open-ended NLG tasks, for which extremum equivalence (\S\ref{sec:limitations})---a condition necessary for aggregating utility across annotators---cannot be satisfied.
    However, magnitude estimation (on a continuous scale) is still possible when using \textit{a single annotator}, since extremum equivalence only applies across annotators.
    
    \item Annotators may not understand the notion of probability or may not read the outputs assigned to them, providing noisy and biased annotations.
    This problem is not unique to SPA, but since human preferences are inherently subjective, identifying insincere annotators is more difficult.
    The filtering strategy of asking annotators to estimate both $P[X \succ_i Y]$ and $P[Y \succ_i X]$ and excluding those for whom $\hat{P}[X \succ_i Y] + \hat{P}[Y \succ_i X] > \tau = 1.1$ proved to be successful in our DALL-E experiments, though there may be even better strategies.
    Also, since we want to estimate the aggregate preference of an agent population, we have to use $n_A$ \underline{unique} agents, instead of letting a few talented annotators do most of the work, as is common in NLP \citep{geva2019we}.
    
    \item There is no simple way to aggregate preference probabilities along multiple axes (e.g., is $X$ more coherent/factual/grammatical than $Y$?).
    Assuming that these factors are independent is not realistic, since one may be downstream of another.
    When doing OAA, the standard practice is to simply take an unweighted average of the factors' Likert scores, but this presumes that equal importance should be given to each factor.
    Under the principles of preference modeling discussed in this paper, practitioners should delegate the task of creating an overall preference to the annotator themself.
    That is, in addition to judging whether $X$ is more coherent/factual/grammatical than $Y$, annotators should also directly judge whether they prefer $X$ to $Y$.
\end{enumerate}

\section*{Ethics Statement}
Accurately reflecting the preferences of users is an ethical imperative when building NLG systems.
Our work can help practitioners be more cognizant of the assumptions and limitations in their evaluation protocol and the broader risks of deploying improperly tested NLG systems.
Our own experiments were conducted with English-speaking US residents, whose preferences are not necessarily representative of the broader population that interfaces with some form of NLG system.

\section*{Acknowledgements}
We thank Kaitlyn Zhou and Tatsunori Hashimoto for feedback on this work. We also thank Clara Meister and Simran Arora for answering queries about annotation on MTurk. KE was supported by a Facebook Fellowship. This work was also supported by NSF award IIS-2128145.

\bibliography{anthology,custom}
\bibliographystyle{acl_natbib}

\appendix

\section{GPT-3 Experiment}
\label{sec:gpt3_appendix}

\subsection{Story Generation}

For each annotator, we first randomly sampled $m = 5$ story prompts from the WritingPrompts datasets after filtering out any prompt that: (1) did not begin with [ WP ]; (2) contained a question mark; (3) did not end in punctuation.
This was done so that the writing prompts were all of a consistent format and style.
We observed that prompts ending in questions sometimes elicited opinion essays from GPT-3, as opposed to a fictional continuation.
In our trial runs, this confused some annotators, who thought all writers were writing fictional continuations.
We thus over-corrected by excluding all prompts with questions.

For each writing prompt, we generated a story by each of the four GPT-3 variants: $\texttt{davinci-002}$, $\texttt{curie-001}$, $\texttt{babbage-001}$, $\texttt{ada-001}$, which we anonymized as writers D,C,B,A respectively.
We set the following hyperparameters for all models: a maximum of 1600 tokens, top-$p$ of 1, and a temperature of 0.9.
Each story prompt came with a human-written continuation as well, which we anonymized as writer E.
In practice, the GPT-3 models usually generated far fewer than the allowable 1600 tokens, resulting in the human-written stories being longer than their machine-written counterparts.
To prevent annotators from using story length as a proxy for quality, we trimmed---to the nearest whole sentence---the human-written story for each prompt so that it was no longer than the longest machine-written story for that prompt.

\begin{figure*}
    \centering
    \includegraphics[width=\textwidth]{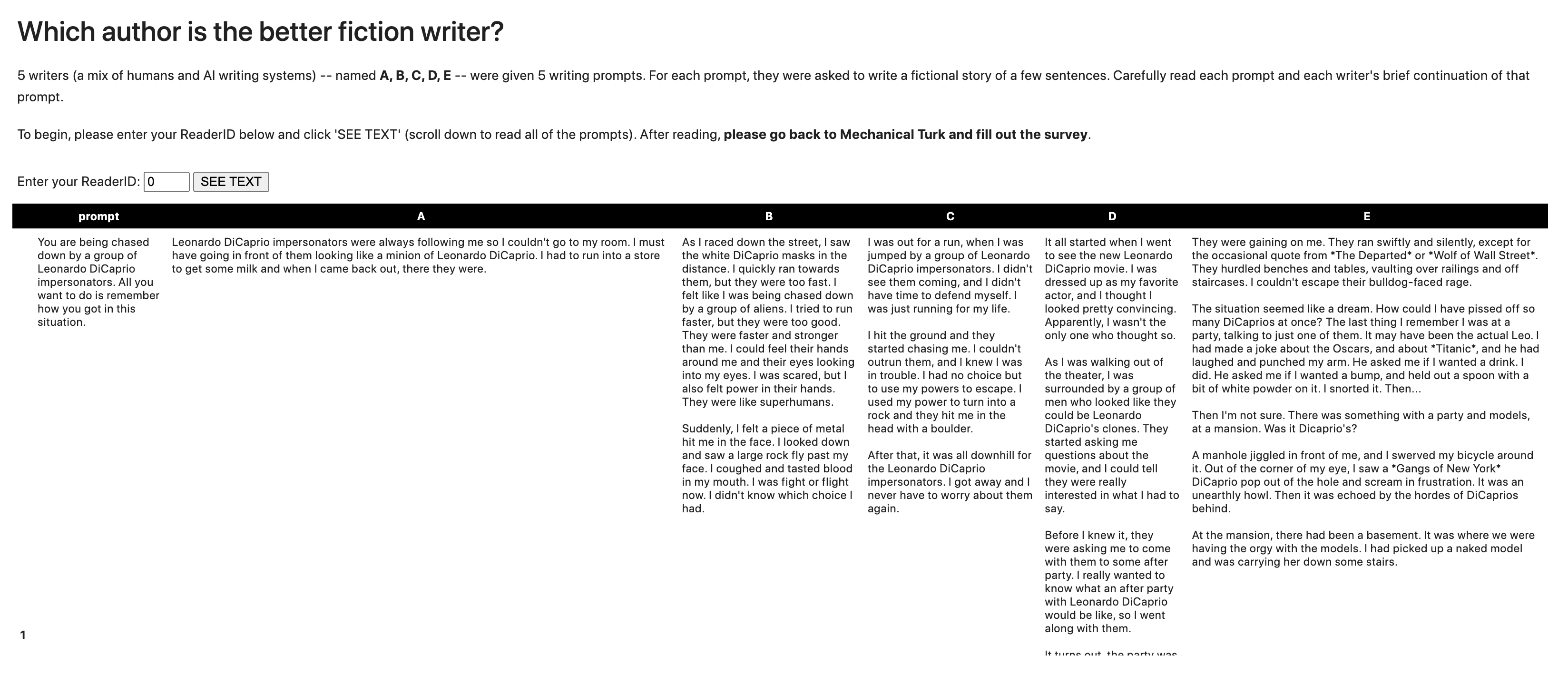}
    \caption{The interface to the generated stories. The continuations generated by the GPT-3 models (A,B,C,D) and the human-written continuation (E) were placed side-by-side.}
    \label{fig:static}
\end{figure*}

The 25 continuations (5 writers $\times$ 5 prompts) that each annotator had to read were put up on a static website, where the annotator would input their assigned ID to read the batch that was assigned to them (see Figure \ref{fig:static}).
The annotator was informed that there were a mix of human and AI writing systems, but we did not reveal which writers were which or how many of each there were.

\begin{figure*}
    \includegraphics[width=\textwidth]{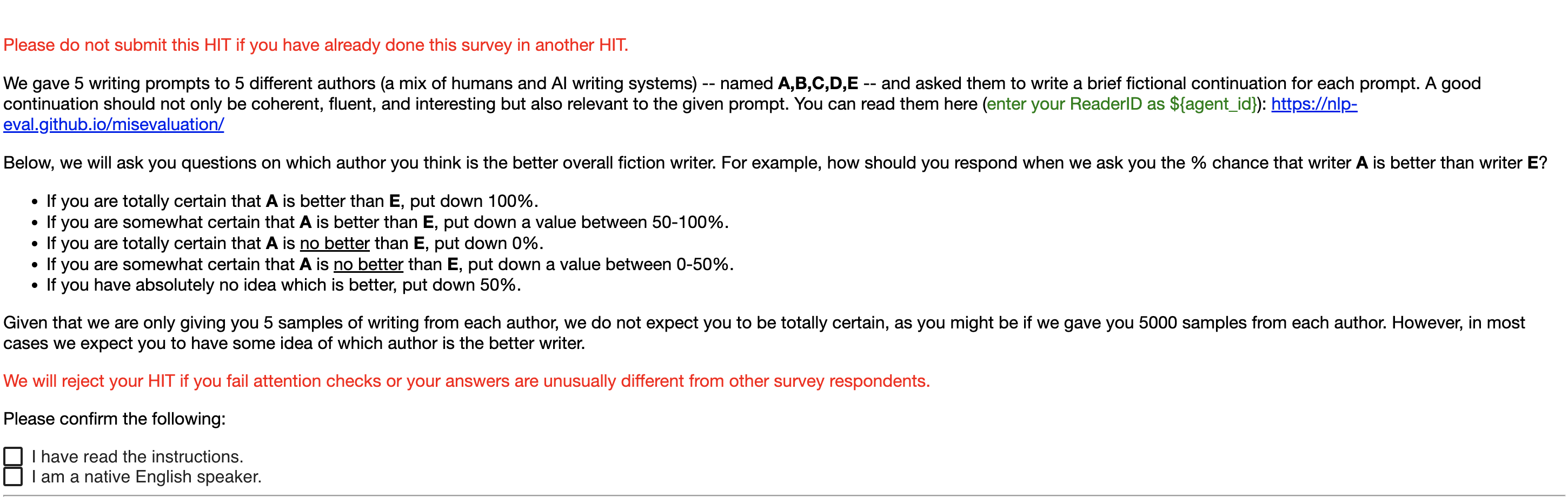}
    \includegraphics[width=0.8\textwidth]{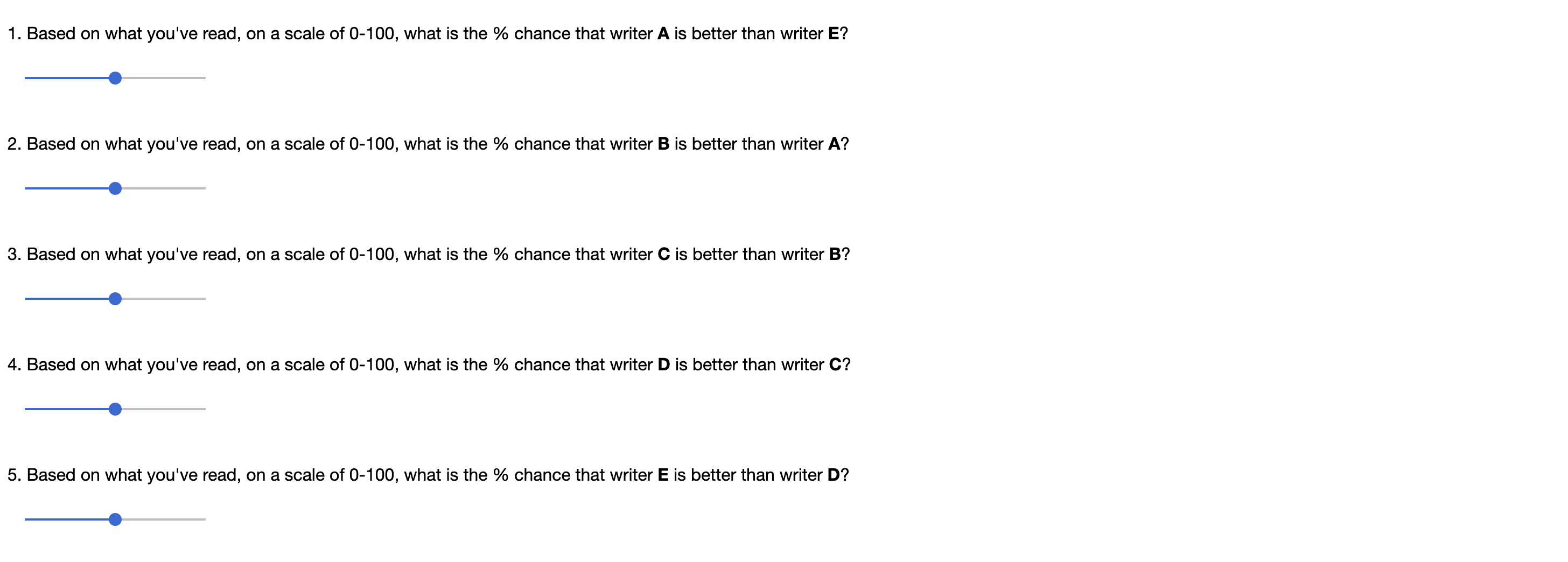}
    \includegraphics[width=\textwidth]{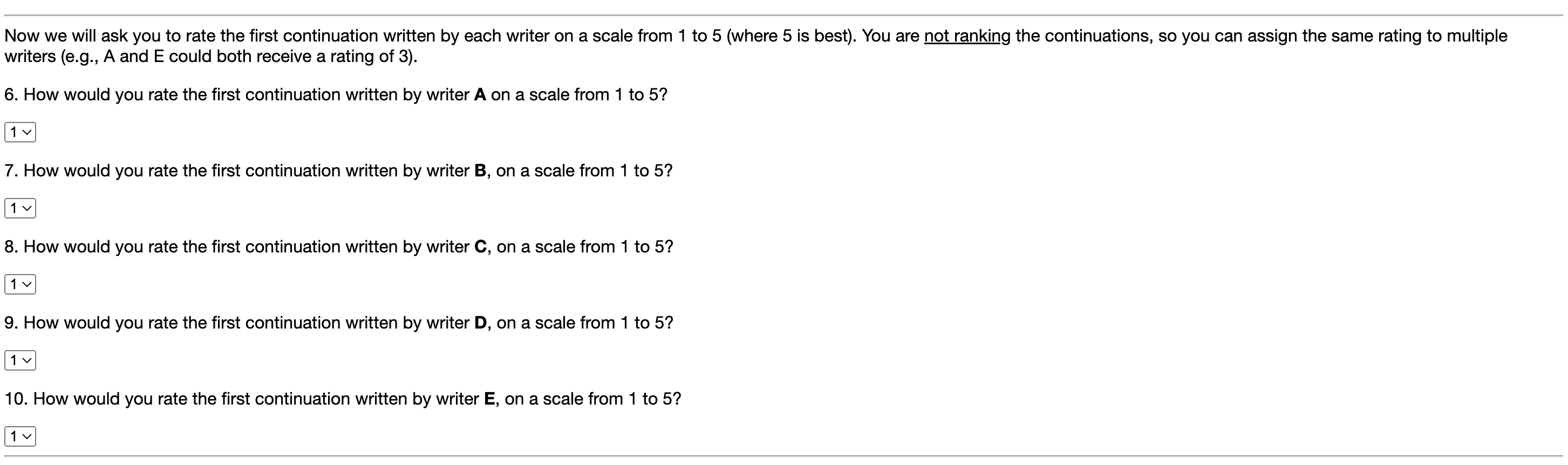}
    \caption{The instructions given to annotators on Amazon Mechanical Turk.}
    \label{fig:mturk}
\end{figure*}

\subsection{Filtering Annotators}
\label{ssec:filtering_ants}

We recruited $n_A = 100$ annotators on Amazon Mechanical Turk, filtering for those who were in the US, had a HIT approval rate $> 98\%$, and who had completed at least 100 HITs.
Each annotator was paid \$5 for approximately 20 minutes of work, working out to roughly USD \$15/hour.
Each annotator was presented with the instructions in Figure \ref{fig:mturk} and then asked to provide 5 preference probabilities $P[X \succ_i Y]$, one for each comparison of interest.
They were asked to evaluate each writing system on the basis of how coherent, fluent, interesting and relevant to the prompt the stories were.

They were then asked to provide a Likert rating of the first continuation written by each writer.
We did not ask for a rating of all 25 continuations because that would have been onerous and unnecessary; for an apples-to-apples comparison of SPA and the standard evaluation protocol, we had an equal sample size for each, giving us 100 probability estimates of the preference and 100 Likert rating deltas that we could feed into a Student's $t$-test.
Since the order of the prompts was random, asking the annotator to provide a Likert rating for the first continuation (as opposed to say, the second or third) made no systematic difference.

After the annotators provided their annotations, we excluded those who: (1) said they were not native English speakers; (2) did not follow our instructions and submitted multiple HITs.
10\% of annotators were excluded, leaving 90 whose annotations we used.
Submitting multiple HITS was an issue because we wanted to control the amount of exposure that the annotator had to each writing system, which is why we provided exactly 5 samples from each.
Annotations were collected in small batches to prevent the same annotators from making multiple submissions.

Note that we did not implement the additional filtering suggested in \S\ref{ssec:new_protocol}, namely excluding annotators for whom $\hat{P}[X \succ_i Y] + \hat{P}[Y \succ_i X] > \tau = 1.1$ for any pair of systems $(X,Y)$ being compared.
This was to see how well SPA would work with minimal annotator filtering.
It is only in the DALL-E experiment (Appendix \ref{sec:dalle_appendix}) that we explored changes to the experiment design, finding that filtering out these self-contradicting annotators is indeed beneficial.

\section{DALL-E Experiment}
\label{sec:dalle_appendix}

\subsection{Image Generation}

We started with 15 image prompts, collected from the DALL-E website itself.
Each prompt was scrambled by randomly re-ordering its tokens.
Both the original prompt and the scrambled prompt were fed to DALL-E, generating 4 images for each.
We call the pipeline that scrambles the prompt before feeding it to DALL-E \texttt{DALL-E-scrambled} and the pipeline that feeds the prompt as is \texttt{DALL-E-raw}.
If the original (unscrambled) prompt is used as the reference, we would expect the latter system to be preferred in aggregate (i.e., $\texttt{DALL-E-raw} \succ \texttt{DALL-E-scrambled}$), since scrambling destroys some compositional concepts.
For example, if the original prompt is \emph{`ball on a chair'}, then an image containing a ball atop a chair is preferable to one that contains a chair atop a ball.

To understand the role that $m$, the number of examples shown, plays in preference probability, we asked the annotator to make four comparisons:
\begin{enumerate}
    \item \textbf{A vs.\ B}, where A is \texttt{DALL-E-raw} and B is \texttt{DALL-E-scrambled}. 
    We randomly sampled (without replacement) \textbf{1} of the 15 original prompts and then uniformly randomly sampled 1 of the 4 original-prompt-based images and 1 of the 4 scrambled-prompt-based images. 
    
    \item \textbf{C vs.\ D}, where C is \texttt{DALL-E-raw} and D is \texttt{DALL-E-scrambled}. 
    We randomly sampled (without replacement) \textbf{2} of the 15 original prompts and then uniformly randomly sampled 1 of the 4 original-prompt-based images and scrambled-prompt-based images for each. 
    
    \item \textbf{E vs.\ F}, where E is \texttt{DALL-E-raw} and F is \texttt{DALL-E-scrambled}. 
    We randomly sampled (without replacement) \textbf{4} of the 15 original prompts and then uniformly randomly sampled 1 of the 4 original-prompt-based images and scrambled-prompt-based images for each.
    
    \item \textbf{G vs.\ H}, where G is \texttt{DALL-E-raw} and H is \texttt{DALL-E-scrambled}. 
    We randomly sampled (without replacement) \textbf{8} of the 15 original prompts and then uniformly randomly sampled 1 of the 4 original-prompt-based images and scrambled-prompt-based images for each.
\end{enumerate}
Thus the annotator was given the impression that they were seeing images from 8 unique image-generation systems.
This concealment is necessary; if the annotator knew that systems C and A were the same, then they may have used the images in the A vs.\ B comparison when judging C vs.\ D, which would have precluded us from studying the effect of $m$.

\subsection{Filtering Annotators}

We recruited $n_A = 60$ annotators on Amazon Mechanical Turk, filtering for those who were in the US, had a HIT approval rate > 98\%, and who had completed at least 100 HITs. 
Each annotator was paid $3$ for approximately 10 minutes of work, working out to roughly USD \$18/hour.
For each comparison of the form $X$ vs.\ $Y$, we asked the annotator to estimate both $P[X \succ_i Y]$ and $P[Y \succ_i X]$.
Since there are four comparisons and two estimates per comparison, a total of eight questions were answered by each annotator.

As in the GPT-3 experiment, we excluded those who said they were not native English speakers or who did not follow our instructions and submitted multiple hits.
We also applied the additional filtering step suggested in \S\ref{ssec:spa_theory} for filtering out annotators with a poor understanding of probability: excluding those for whom $\hat{P}[X \succ_i Y] + \hat{P}[Y \succ_i X] > \tau = 1.1$ for any $(X,Y) \in \{(A,B), (C,D), (E,F), (G,H)\}$.
The filtering left 29 eligible annotators.

\begin{figure}
    \centering
    \includegraphics[width=\columnwidth]{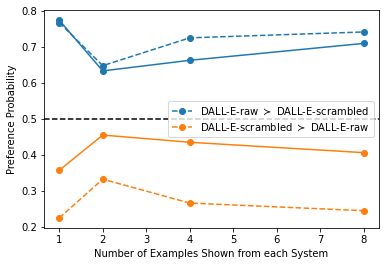}
    \caption{Preferences probabilities for the DALL-E image-generation experiment before filtering annotators (solid) and after filtering (dashed).
    Though all preference probabilities are in the right direction, filtering out annotators with a poor understanding of probability increased effect sizes and restores the symmetry that should exist under Assumption 2 (i.e., $\hat{P}[X \succ_i Y] \approx 1 - \hat{P}[Y \succ_i X]$).
    }
    \label{fig:filter}
\end{figure}

As seen in Figure \ref{fig:filter}, the benefits of this additional filtering step were two-fold:
\begin{enumerate}
    \item The remaining annotators are more certain about their preference, leading to a larger effect size.
    \item The excluded annotators are less willing to commit to a preference than a dispreference (e.g., less willing say that $\hat{P}[Y \succ_i X] < 0.5$ than say that $\hat{P}[X \succ_i Y] > 0.5$, though both are semantically equivalent).
\end{enumerate}
For these reasons, we recommend practitioners follow the optional filtering step suggested in \S\ref{ssec:new_protocol}.

\end{document}